\documentclass[10pt,twocolumn,letterpaper]{article}

\usepackage[pagenumbers]{cvpr} % To force page numbers, e.g. for an arXiv version

\usepackage{multirow}
\definecolor{cvprblue}{rgb}{0.21,0.49,0.74}
\usepackage[pagebackref,breaklinks,colorlinks,allcolors=cvprblue]{hyperref}

\def\paperID{6345} % *** Enter the Paper ID here
\def\confName{CVPR}
\def\confYear{2026}

\title{Pseudo-Stereo Inputs: A Solution to the Occlusion Challenge in Self-Supervised Stereo Matching}

\author{
    Ruizhi Yang\textsuperscript{\rm1,\rm2,\rm3,\rm4} \quad Xingqiang Li\textsuperscript{\rm1,\rm2,\rm4,\thanks{Corresponding authors}} \quad Jiajun Bai\textsuperscript{\rm1,\rm2,\rm4} \quad Jinsong Du\textsuperscript{\rm1,\rm2,\rm4}\\
    \textsuperscript{\rm1}Shenyang Institute of Automation, Chinese Academy of Sciences \quad \textsuperscript{\rm2}Liaoning Liaohe Laboratory \\
    \textsuperscript{\rm3}University of Chinese Academy of Sciences \\
    \textsuperscript{\rm4}Key Laboratory on Intelligent Detection and Equipment Technology of Liaoning Province \\
    \tt\small \{yangruizhi, \tt\small lixingqiang, \tt\small\ baijiajun\}@sia.cn \quad \tt\small jsdu\_sia@163.com
}

\begin{document}
\maketitle
\begin{abstract}
Self-supervised stereo matching holds great promise by eliminating the reliance on expensive ground-truth data. Its dominant paradigm, based on photometric consistency, is however fundamentally hindered by the occlusion challenge---an issue that persists regardless of network architecture. The essential insight is that for any occluders, valid feedback signals can only be derived from the unoccluded areas on one side of the occluder. Existing methods attempt to address this by focusing on the erroneous feedback from the other side, either by identifying and removing it, or by introducing additional regularities for correction on that basis. Nevertheless, these approaches have failed to provide a complete solution.
This work proposes a more fundamental solution. The core idea is to transform the fixed state of one-sided valid and one-sided erroneous signals into a probabilistic acquisition of valid feedback from both sides of an occluder. This is achieved through a complete framework, centered on a pseudo-stereo inputs strategy that decouples the input and feedback, without introducing any additional constraints. Qualitative results visually demonstrate that the occlusion problem is resolved, manifested by fully symmetrical and identical performance on both flanks of occluding objects. Quantitative experiments thoroughly validate the significant performance improvements resulting from solving the occlusion challenge. %The code is provided in the supplementary materials and will be publicly available upon receipt.
\end{abstract}    
\section{Introduction}
\label{sec:intro}

Stereo matching, a cost-effective method for obtaining reliable 3D information, has been a longstanding focus of research in computer vision \cite{poggiSynergiesMachineLearning2021}. In recent years, learning-based stereo matching methods have emerged, demonstrating superior performance in terms of accuracy and efficiency \cite{mayerLargeDatasetTrain2016,changPyramidStereoMatching2018,li2024local}. Nevertheless, the dominant paradigm remains fully reliant on ground-truth annotations, fundamentally limiting its potential. Self-supervised stereo matching \cite{zhouUnsupervisedLearningStereo2017,liOcclusionAwareStereo2019,wangUnifiedUnsupervisedOpticalFlow2019a,wangParallaxAttentionUnsupervised2022} was introduced to resolve this annotation issue and initially progressed in tandem with supervised methods, but its development soon reached a bottleneck, hindering further advancement.

The guiding principle of self-supervised stereo matching is the photometric consistency  between different views \cite{godardUnsupervisedMonocularDepth2017}. This principle, however, inherently fails in regions where the consistency assumption is violated, resulting in erroneous feedback signals. Occlusion are the most critical challenge, as they produce large-scale, persistent erroneous feedback that severely degrades network performance. Consequently, a significant body of work has focused on addressing this problem \cite{godardUnsupervisedMonocularDepth2017,liOcclusionAwareStereo2019,fanOcclusionAwareSelfSupervisedStereo2022,liUnsupervisedOcclusionawareStereo2022,yangUnsupervisedHierarchicalIterative2024}, including related domains facing the same challenge \cite{Meister:2018:UUL,wangOcclusionAwareUnsupervised2018,jonschkowski2020matters,hurSelfSupervisedMultiFrameMonocular2021}. Their approaches can be broadly summarized into a two-pronged strategy: one is to identify occluded regions to prevent the backpropagation of erroneous gradients, and the other is to compensate for the missing information in these areas by introducing additional sources of feedback.

\begin{figure}
    \centering
    \includegraphics[width=1\linewidth]{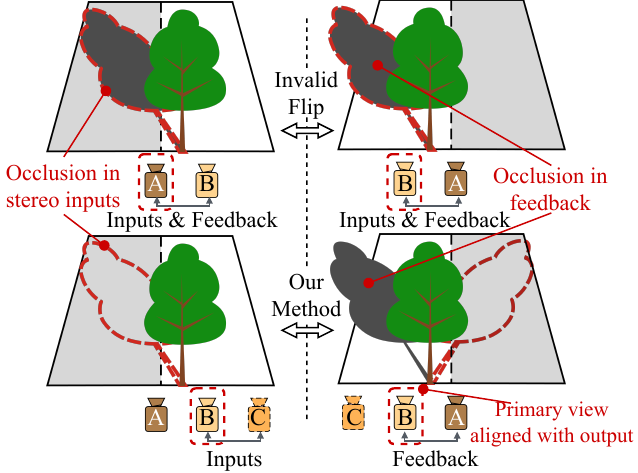}
    \caption{An illustration of our motivation. (Top) The fixed-side occlusion problem where standard flipping is ineffective, as the occlusion side remains constant relative to the inputs. (Bottom) Our method decouples inputs and feedback using a pseudo-view~(C), breaking the persistent directional error signal.}
    \label{fig:core}
\end{figure}

Information restoration for occluded areas remains a major challenge, and existing methods have yet to achieve satisfactory results. Mainstream approaches rely on providing auxiliary feedback through mechanisms like an edge-aware smoothness loss \cite{godardUnsupervisedMonocularDepth2017,jonschkowski2020matters,wangParallaxAttentionUnsupervised2022}, while others attempt to exploit local disparity regularities \cite{liUnsupervisedOcclusionawareStereo2022,yangUnsupervisedHierarchicalIterative2024}. Fundamentally, these methods depend on information from the immediate neighborhood, which presumes a high degree of regularity and correlation between the occluded region and its surroundings—an assumption that is often unreliable. Some works \cite{godardUnsupervisedMonocularDepth2017,liOcclusionAwareStereo2019,fangESNetAccurate2023} advocate for using a flip data augmentation to source information from symmetric regions. However, as demonstrated in \cref{fig:core}, this strategy is ineffective. Due to the coupled nature and fixed directionality of the input and feedback signal, occlusions consistently appear on a fixed side of the occluding object in the primary view, and the flip operation fails to alter this geometric reality.
In contrast, methods \cite{yuanStereoMatchingSelfsupervision2021,tosiNeRFSupervisedDeepStereo2023} that introduce a third viewpoint—using a central primary view and selecting the minimum loss from the left and right views—have proven effective at correctly resolving occlusions. Nevertheless, these solutions are impractical for standard settings. They either necessitate specialized, perfectly symmetric trinocular camera hardware \cite{yuanStereoMatchingSelfsupervision2021} or require a full 3D scene reconstruction via methods like NeRF to synthesize the novel view \cite{tosiNeRFSupervisedDeepStereo2023}. Consequently, such techniques cannot be directly applied for self-supervised training on conventional binocular image data.

In this work, we propose a novel occlusion handling method that acquires reliable feedback, instead of relying on simple inference from local neighbors, and also without introducing any additional constraints. The core idea, illustrated in \cref{fig:core}, is to decouple the network input and its photometric feedback signal by using a pseudo third-view input. This principle effectively reframes the problem from one of static, one-sided occlusions to one of dynamic, two-sided occlusions with equal probability. The execution of this idea is non-trivial, and we identified and solved three core challenges. First, how to generate the pseudo third-view input with minimal cost and using it to achieve the aforementioned transformation to symmetric occlusions. Second, given that symmetric occlusions imply that each side of an occluder receives correct and incorrect feedback with equal probability, how to guide the network to converge to the correct solution. Third, how to ensure proper network convergence and prevent overfitting, given the distinguishable nature of the pseudo-input compared to the real input.

Our experiments validate that the proposed full framework reliably resolves the occlusion problem, manifested by the symmetric and accurate estimates on both sides of occluders shown in \cref{fig:kt}, and the significant performance gains reported in \cref{tab:online}. Furthermore, subsequent experiments (\cref{fig:more_datasets} and \cref{tab:more_datasets}) confirm that occlusion is a core bottleneck in self-supervised stereo matching, independent of network architecture or dataset. We demonstrate that our proposed method successfully overcomes this fundamental bottleneck across diverse network architectures and datasets.

In summary, the main contributions of this work are as follows.
\begin{enumerate}
    \item We propose a novel occlusion handling strategy that uses pseudo-stereo inputs to decouple network input and feedback. This fundamentally reframes the learning problem in occluded regions.
    \item We introduce a complete and robust framework to realize this idea, ensuring convergence under feedback conflicts and preventing overfitting to synthetic artifacts.
    \item Extensive experiments show our method resolves the occlusion bottleneck to achieve significant performance improvements, generalizing effectively across various network backbones and datasets.
\end{enumerate}
\section{Related Work}
\begin{figure*}
    \centering
    \includegraphics[width=0.85\linewidth]{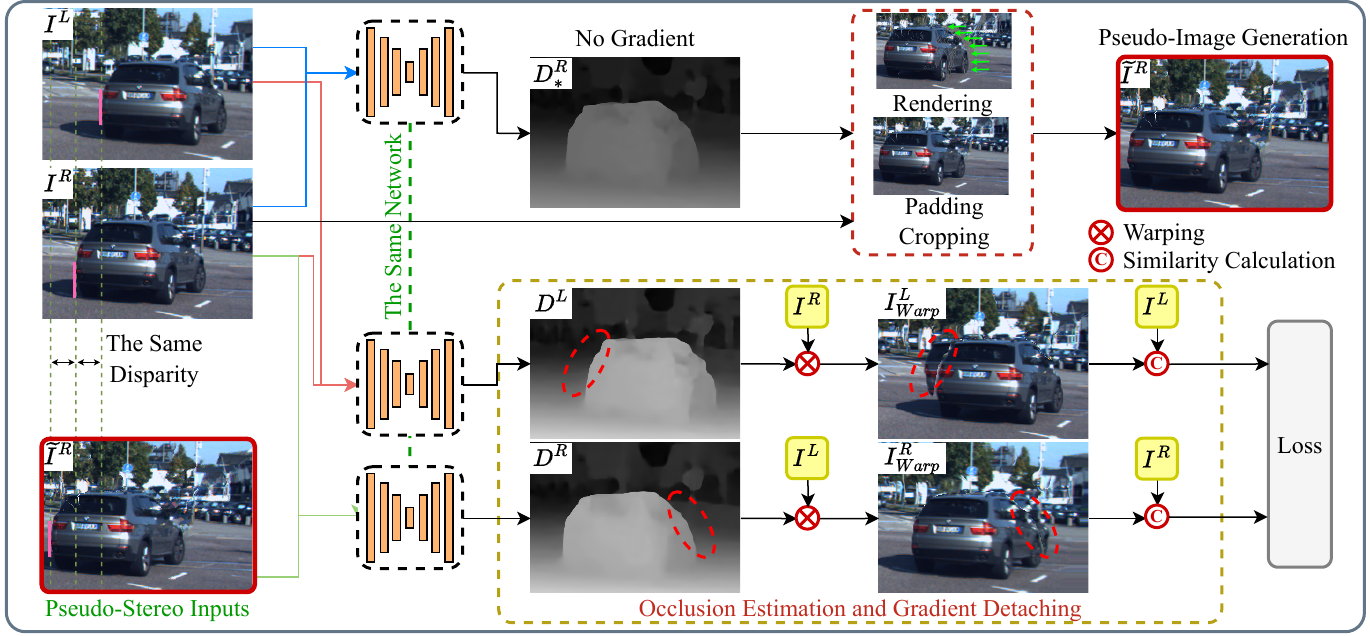}
    \caption{An overview of our proposed pipeline. We leverage a pseudo-input to decouple network inputs and feedback. This enables a symmetric training paradigm where occlusions are probabilistically relocated, providing valid feedback for both sides of occluders. An occlusion estimation and gradient detaching mechanism is employed to resolve the inherent feedback conflicts.}
    \label{fig:overview}
\end{figure*}

\subsection{Learning-Based Stereo Matching Backbone}
The rapid advancement of deep learning has given rise to powerful stereo matching architectures~\cite{hamidStereoMatchingAlgorithm2022,tosiSurveyDeepStereo2025}. Representative works like PSMNet \cite{changPyramidStereoMatching2018} introduced 3D convolutions for cost aggregation, while more recent methods like GMStereo \cite{xu2023unifying} incorporate transformers \cite{transformer} for more capable feature matching. Recently, approaches such as MonSter~\cite{chengMonSterMarryMonodepth2025} have shown success by integrating features from pre-trained monocular depth estimators. Despite their increasingly powerful feature representation and matching capabilities, these models are fundamentally designed for a supervised paradigm and require accurate ground-truth disparities. When applied to self-supervised learning, where the supervision signal can be ambiguous or erroneous, their advanced feature induction capabilities, while still beneficial, do not address the core bottleneck. Although some works \cite{liOcclusionAwareStereo2019,wangParallaxAttentionUnsupervised2022,fanOcclusionAwareSelfSupervisedStereo2022} have proposed architectural modifications specifically for self-supervised occlusion handling, their visual results reveal that the underlying problem is not significantly mitigated. This suggests that the solution to the occlusion problem lies primarily within the loss formulation rather than the network architecture itself.

\subsection{Self-Supervised Stereo and Occlusion Handling}
Self-supervised stereo methods are built upon the principle of photometric consistency between views \cite{zhouUnsupervisedLearningStereo2017,wangUnifiedUnsupervisedOpticalFlow2019a,wangParallaxAttentionUnsupervised2022,yangUnsupervisedHierarchicalIterative2024, fangESNetAccurate2023}. Consequently, a large body of work has focused on strategies to address this principle's primary weakness: the erroneous feedback generated by occlusions. A common consensus is to identify occluded regions via a mask and exclude them from the photometric loss, typically through pixel-wise multiplication \cite{Meister:2018:UUL,wangOcclusionAwareUnsupervised2018,wangUnifiedUnsupervisedOpticalFlow2019a,jonschkowski2020matters,wangParallaxAttentionUnsupervised2022}. Method \cite{jonschkowski2020matters} compared various occlusion inference methods \cite{wangOcclusionAwareUnsupervised2018,Meister:2018:UUL,Janai_2018_ECCV}, demonstrating that all benefit performance, and crucially established that the gradient stop for occlusion mask is an important strategy.

With occlusions masked, however, an information void is created. Various methods have been proposed to compensate for this. Without introducing external constraints, this compensation is typically provided by a standard smoothness loss \cite{wangParallaxAttentionUnsupervised2022,Meister:2018:UUL,wangOcclusionAwareUnsupervised2018,wangUnifiedUnsupervisedOpticalFlow2019a,jonschkowski2020matters}. \cite{jonschkowski2020matters} demonstrated that different smoothness edge-weights can significantly impact performance. Nevertheless, since smoothness is not always a reliable or correct convergence target, this approach remains unsatisfactory. Some methods \cite{godardUnsupervisedMonocularDepth2017,liOcclusionAwareStereo2019,fangESNetAccurate2023} employ flip strategy and regard left-right consistency directly as a loss term, attempting to use it as a key solution. However, this does not fundamentally mitigate the occlusion problem, as an incorrect disparity can still satisfy the consistency. A more targeted line of work \cite{liUnsupervisedOcclusionawareStereo2022,yangUnsupervisedHierarchicalIterative2024} have designed particular losses to extract more reliable information from local neighborhoods. Among these, \cite{liUnsupervisedOcclusionawareStereo2022} focuses on interpolating from background neighbors rather than foreground ones, while \cite{yangUnsupervisedHierarchicalIterative2024} encourages occluded regions and their neighborhoods to conform to a 3D geometry inferred from monocular cues. These methods are more tailored for occlusion handling and have achieved performance gains, but their solutions are not robust as they rely on the strong assumption that neighboring information is sufficiently correlated with the occluded content.

Distinct from the aforementioned methods, some works \cite{yuanStereoMatchingSelfsupervision2021,tosiNeRFSupervisedDeepStereo2023} acquire reliable, non-local feedback by leveraging additional views. Specifically, \cite{yuanStereoMatchingSelfsupervision2021} uses a fixed, symmetric multi-view camera hardware where a region occluded from one direction is visible from another. \cite{tosiNeRFSupervisedDeepStereo2023} achieves the same goal by rendering a third-view image using Neural Radiance Fields \cite{mildenhall2020nerf} (NeRF). These two approaches provide a conceptually correct pathway for occlusion handling: obtaining unoccluded information from an auxiliary perspective. However, both methods either depend on specialized hardware or require additional data for scene reconstruction, imposing external constraints that prevent their direct application to conventional binocular datasets. This is inconsistent with the goal of self-supervised learning, which aims for direct training on any target scene. 

In this paper, we propose a self-supervised method that can acquire the same reliable feedback as these multi-view approaches without depending on any such external constraints.

\section{Method} \label{sec:method}

The motivation and the overall framework of our method are illustrated in \cref{fig:core} and \cref{fig:overview}, respectively. Section \ref{sub-pseudo} details the core idea in our approach: a pseudo-stereo inputs strategy that decouples input images from feedback images. Section \ref{sub-feedback} further discusses the practical challenges of this design and proposes solutions, including addressing feedback conflicts and overfitting problems.

\subsection{Pseudo-Stereo Inputs Strategy} \label{sub-pseudo}
This section elucidates the principles and implementation of our pseudo-stereo inputs strategy, which addresses the occlusion challenge by decoupling the input image pairs from the feedback image pairs in self-supervised training. The core method for this decoupling is to utilize image pairs containing a generated pseudo-image as inputs instead of original image pairs. In contrast, the feedback calculation relies solely on the original images to ensure the reliability and accuracy of the feedback. This decoupling removes the fixed relative positional constraints on the input and feedback images, allowing them to flexibly select positions where occlusions occur, thereby addressing the issue of continuous information loss due to fixed occlusions. An intuitive illustration of the principle is shown in \cref{fig:core}.%The following subsections provide a detailed discussion. %Figure xx illustrates the generation of a pseudo-image utilizing the current network's predicted disparity map and the original view. 

\begin{figure}
    \centering
    \includegraphics[width=1\linewidth]{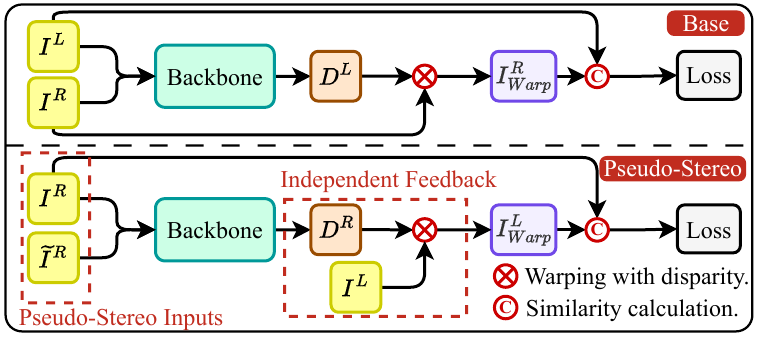}
    \caption{A comparison between the existing self-supervised pipeline and our pseudo-stereo inputs strategy. In contrast to existing methods, our method decouples the input stereo images from the feedback stereo images. This allows for flexible feedback using image pairs with different occlusion positions.}
    \label{fig:pipeline}
\end{figure}

\subsubsection{Pseudo-Stereo Inputs}
Figure \ref{fig:pipeline} visually illustrates the distinction between this strategy and existing self-supervised paradigms. In this strategy, the left view remains a real image from the dataset, while the right view is replaced with a generated pseudo-image. Specifically, existing methods use a stereo image pair $I^L$,$I^R$ from the dataset as network inputs and output a disparity map $D^L$ aligned with $I^L$. Where $I$ and $D$ denote color images and disparity maps, respectively. Superscripts $L$ and $R$ indicate correspondence to the original left and right viewpoints. In contrast, when using pseudo-inputs, an estimated disparity $D^R$ and $I^R$ are used to generate a pseudo-image $\widetilde{I}^R$. At this point, the network inputs become $I^R$ as the left view and $\widetilde{I}^R$ as the right view, producing a dense disparity map $D^R$ aligned with $I^R$. The images used for feedback should be the original ones to ensure reliability. Therefore, the original image $I^L$ is warped using $D^R$ to obtain $I_{Warp}^L$, which is aligned with $I^R$. $I^R$ and $I_{Warp}^L$ are then employed for loss calculation. When $I^L$ is as the left input, the occluded areas appear to the left of the occluders; conversely, when $I^R$ is as the left input, the occluded areas are on the right side of the occluders. This strategy ensures that the network can consistently have a probability to capture information from both sides of the occluders, thus resolving the problem where previous methods invariably lacked information from one side of the occluders. 

\subsubsection{Method for Generating Pseudo-Images} \label{sub:gen}
\begin{figure}
    \centering
    \includegraphics[width=\linewidth]{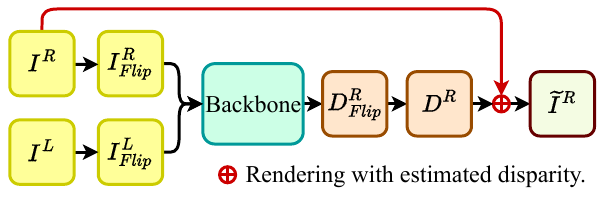}
    \caption{Schematic diagram of the pseudo-images generation using the rendering method, where Backbone is the model currently under training rather than a previously well-trained one. The subscript ``Flip'' denotes horizontal flipping.}
    \label{fig:gen}
\end{figure}

While numerous methods for novel view synthesis exist \cite{khanTiledMultiplaneImages2023,mildenhall2020nerf,kerbl3DGaussianSplatting2023}, they typically require additional data to train the models themselves. Incorporating such methods would introduce extraneous data and priors, which contradicts the fundamental principle of self-supervised learning. Therefore, our approach must exclusively utilize information available within the original training pipeline. In the standard stereo estimation pipeline, the commonly used warping operation is a form of image synthesis; this involves using $D^L$, which is aligned with $I^L$, to warp $I^R$ into an novel image that aligns with $I^L$. However, this approach cannot be used for generating pseudo-images. The reason is that we do not have a disparity map aligned with the pseudo-image prior to its generation. Consequently, our approach employs a pseudo-image generation method that more closely resembles a simplified rendering process as illustrated in \cref{fig:gen}. For $I^R$, each pixel can be relocated backward using $D^R$ to generate $\widetilde{I}^R$. If multiple pixels map to the same location in $\widetilde{I}^R$, only the pixel with the maximum disparity is kept. This corresponds to the occlusion effect during rendering. As shown in \cref{fig:rendering}, this process yields a pseudo-image containing missing pixels. To make it as similar to the real image as possible, it is necessary to pad these missing pixels. The padding method and how closely the resulting image resembles a real one directly influence the severity of the overfitting problem. This issue is discussed in detail in \cref{sub-feedback}.

\subsubsection{Discussion on Using Pseudo Images as Inputs}
As described earlier, our method does not use a previously trained network to generate novel views. Instead, the generation of pseudo-images relies on the network that is currently undergoing training. In other words, images generated in the early stages of training, which severely lack realism, are also used as inputs during the training process. This raises concerns about the network's convergence. However, in fact, forcing the network to exploit monocular cues during stereo training is a common and effective practice.
As an example, in the supervised stereo matching method \cite{tankovichHITNetHierarchicalIterative2021}, random patches are used to occlude the right view to accomplish this objective. 

The cornerstone of our method's stable convergence is that the left input and the feedback image are always real and reliable. The effective feedback compels the network to leverage monocular cues in the initial stages of training to enhance prediction accuracy. As accuracy increases, the precision of pseudo-image generation also improves, leading to better binocular information for the network. This results in a stable and progressively enhancing iterative process. The experimental results in Section \ref{sub-ablation} also support this point. In these experiments, even when no real images were used for the right input (i.e., using $I^L$, $\widetilde{I}^L$ or $I^R$, $\widetilde{I}^R$ as network inputs), the network still converges successfully.

\subsection{Feedback Conflicts and Overfitting Problems} \label{sub-feedback}
Simply applying the pseudo-stereo inputs strategy with no further modifications already enables a significant performance improvement over the standard paradigm, which can be evident from the ablation study in \cref{sub-ablation}. However, the feedback conflicts and overfitting issues hinder further performance enhancements and prevent stable convergence.

\subsubsection{Feedback Conflicts}
The feedback conflicts issue manifests as unstable disparity estimates in occluded regions. This phenomenon can be readily explained. Despite our strategy in self-supervised feedback ensuring a 50\% probability of obtaining correct information on the occluded side, there remains an equal 50\% probability of receiving incorrect information. Under the influence of this conflicting feedback, the network will attempt to minimize the loss of both correct and incorrect information simultaneously, finally resulting in a compromised outcome in the occluded region.
More importantly, the gradient of the photometric loss is negatively correlated with image similarity. As training progresses and overall image similarity increases, the global average gradient diminishes to a low value. However, occluded regions, which lack correct correspondences, produce a large color discrepancy when warped even with the correct disparity.

Occlusion estimation and detaching the propagation of loss from occluded pixels offers a promising solution to this challenge. While consistency check \cite{hirschmullerAccurateEfficientStereo2005} is an established method for occlusion estimation, it suffers from two notable drawbacks. First, it incurs additional computational cost by requiring the inference of a disparity map for the opposing view. Second, it is prone to errors where incorrect disparities coincidentally satisfy the consistency check, a problem particularly prevalent at object boundaries. We implement a solution that leverages the inherent properties of the current view rather than relying on information from another view. Specifically, we identify invalid pixels during the rendering process---those falling outside the image boundaries or occluded by other pixels---and classify them as occlusion pixels. This method effectively minimizes the risk of misclassifying pixels and prevents erroneous feedback from propagating. After mitigating erroneous feedback stemming from occluded pixels using this approach, we observe a notable improvement in the network's performance. Detailed results can be found in the ablation study presented in \cref{sub-ablation}. Subsequently, the photometric loss $L_p$ can be expressed as:
\begin{equation} \label{eq:lp1}
    L_p = \frac{1}{N}\sum_{ij}{(p \cdot pe(I^L_{ij}, I^R_{ij}) + (1 - p) \cdot pe(I^R_{ij}, \widetilde{I}^R_{ij})) \cdot \widetilde{O}_{ij}}
\end{equation}
\begin{equation} 
\begin{aligned}
\text{where } pe(I^A, I^B) &= \frac{\alpha}{2} (1 - SSIM(I^A, I^B)) \\
&+ (1 - \alpha) \|I^A - I^B\|_1 
\end{aligned} 
\end{equation}

Here, $p$ and $1-p$ represent the probabilities of the respective items, both being constant at $0.5$. $\widetilde{O}_{ij}$ denotes the occlusion estimation value, where it equals 1 for non-occluded pixels and 0 for occluded ones. $\alpha=0.85$ as in \cite{godardDiggingSelfSupervisedMonocular2019}.
\subsubsection{Overfitting Problems}
Incorporating occlusion estimation into the pseudo-stereo inputs strategy further enhances performance. Nevertheless, extended long-term training stability tests have revealed the overfitting problem, manifested as a gradual decline in accuracy after reaching a peak. Critically, self-supervised training, by its nature, operates without labeled data and thus lacks a validation set. Consequently, early stopping cannot be relied upon to prevent overfitting; instead, the network must be guaranteed to converge stably.

We identify the root cause of this overfitting as the network's ability to distinguish between real and pseudo images in the later stages of training. Specifically, in current strategies, the use of real or pseudo image pairs directly determines the location of occlusions. This enables the network to classify inputs by detecting whether they are pseudo images, thereby converging different results for each input type. This renders the network no longer ``consistently have a probability to capture information from both sides of the occluders'' as expected.
\begin{figure}
    \centering
    \includegraphics[width=1\linewidth]{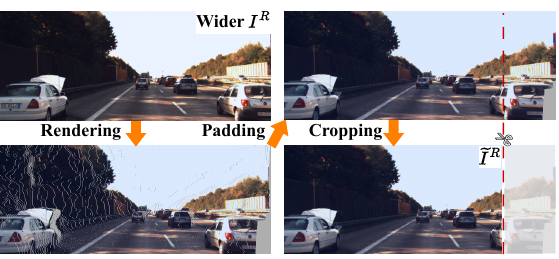}
    \caption{Visual demonstration of the pseudo-image generation process. Issues with small-scale pixel loss and large-scale pixel missing at edges are mitigated using padding and cropping.}
    \label{fig:rendering}
\end{figure}

These observations suggest that the generated pseudo-images contain sufficient features for the network to distinguish them. Based on this inference, we initially attempted to generate more indistinguishable images to prevent the network from classifying the inputs. When generating images using the rendering method described in \cref{sub:gen}, notable features include small-scale voids at occluded areas and large-scale missing areas along the right edge, as shown in \cref{fig:rendering}. For small-scale voids, we fill them using the average value of adjacent non-void pixels. For the missing parts at the right edge, we adopt a wider generation strategy to produce images that are wider than the inputs. This ensures that the resulting cropped images no longer suffer from large-scale missing regions at their edges. It is evident that the generated images now possess a higher degree of realism. Under this strategy, experiments demonstrate an improvement in overfitting issues; however, they have not been completely eradicated. This implies that the network is still able to extract subtle features for differentiation.

We further implement the fully pseudo-stereo inputs strategy. This strategy no longer relies on real image pairs but instead uses fully pseudo-stereo images as input. To be specific, the network's inputs are changed to image pairs $I^L$,$\widetilde{I}^L$ and $I^R$, $\widetilde{I}^R$. At this point, since all the inputs are pseudo-stereo inputs, the network naturally cannot differentiate them further. As shown by the experiments in \cref{sub-ablation}, the network achieves stable convergence to high performance and avoids overfitting under this strategy. At this point, (\ref{eq:lp1}) becomes,
\begin{equation}
    L_p = \frac{1}{N}\sum_{ij}{(p \cdot pe(I^L_{ij}, \widetilde{I}^L_{ij}) + (1 - p) \cdot pe(I^R_{ij}, \widetilde{I}^R_{ij})) \cdot \widetilde{O}_{ij}}
\end{equation}
In addition to $L_p$, the same edge-aware smoothness $L_s$ as in \cite{godardDiggingSelfSupervisedMonocular2019} is adopted, and the disparity values are also mean-normalized like them.
\begin{equation}
    L_s = |\partial_x D^{L^*}| e^{-|\partial_x I^L|} + |\partial_y D^{L^*}| e^{-|\partial_y I^L|},
\end{equation}
where $D^{L^*} = D^L / \overline{D^L}$. In our experiments, it is observed that an overly large smoothing loss at the early stages hinders the model's exploration capability and has a slight negative impact on its final performance. Therefore, we modify the smoothing term to a parameter that gradually reaches the set value, reducing its influence in the early stages. The final loss can be expressed as:
\begin{equation}
    L = L_p + \lambda L_s. 
\end{equation}
$\lambda$ is the smoothing term coefficient, starting from 0.001 and gradually increasing to 0.5 over 10k iterations.

\begin{table*}[ht]
\centering
\caption{Comparison with existing methods on the KITTI online benchmark.}
\label{tab:online}
\resizebox{2.\columnwidth}{!}{%
\begin{tabular}{lccccccccc}
\toprule
\multirow{2}{*}{Method}                                     & \multicolumn{4}{c}{KITTI 2012 Test}                         &           & \multicolumn{4}{c}{KITTI 2015 Test}                            \\ \cline{2-5} \cline{7-10} 
                                                            & 3-noc (\%)    & 3-all (\%)    & EPE-noc      & EPE-all      &           & D1-bg (\%)    & D1-fg (\%)     & D1-all (\%)   & D1-noc (\%)   \\ \midrule
OASM \cite{liOcclusionAwareStereo2019}                      & 6.39          & 8.60          & 1.3          & 2.0          &           & 6.89          & 19.42          & 8.98          & 7.39          \\
PASMnet \cite{wangParallaxAttentionUnsupervised2022}        & -             & -             & -            & -            &           & 5.41          & 16.36          & 7.23          & 6.69          \\
OASM-DDS \cite{liUnsupervisedOcclusionawareStereo2022}      & 4.81          & 5.69          & 1.1          & 1.2          &           & 4.46          & 15.76          & 6.51          & 6.02          \\
UHP \cite{yangUnsupervisedHierarchicalIterative2024}        & 6.05          & 7.09          & 1.2          & 1.3          &           & 5.00          & 13.7           & 6.45          & 5.93          \\
Flow2Stereo \cite{Flow2StereoEffectiveSelfSupervised}       & 4.58          & 5.11          & 1.0          & 1.1          &           & 5.01          & 14.62          & 6.61          & 6.29          \\
CRD-Fusion \cite{fanOcclusionAwareSelfSupervisedStereo2022} & 4.38          & 5.40          & 0.9          & 1.1          &           & 4.59          & 13.68          & 6.11          & 5.69          \\
Pseudo-Stereo-PSMNet (ours)                                 & 3.46          & 4.08          & 0.8          & 0.9          &           & 3.11          & 12.52          & 4.68          & 4.40          \\
Pseudo-Stereo-MonSter (ours)                                & \textbf{3.08} & \textbf{3.62} & \textbf{0.7} & \textbf{0.8} & \textbf{} & \textbf{2.93} & \textbf{11.67} & \textbf{4.39} & \textbf{4.22} \\ \bottomrule
\end{tabular}%
}
\end{table*}
\section{Experiments}
In this section, we present experiments to validate that our method successfully resolves the occlusion problem. The logic of our experimental validation is as follows. First, in \cref{sub-compar}, we conduct a direct comparison against existing methods on the mainstream benchmark datasets they established. Then, we perform additional experiments across various datasets and backbone networks to verify the generalizability of our method in \cref{sub-more}. Finally, \cref{sub-ablation} presents ablation studies to validate the contribution of each proposed component. We emphasize the visualization of occluded regions to confirm the resolution of the occlusion problem, while also showcasing the performance improvements achieved by solving it.

\textbf{Datasets and Backbones.} For evaluation, we use the KITTI 2012 \cite{geigerAreWeReady2012} and KITTI 2015 \cite{menzeObjectSceneFlow2015} datasets, which consist of real-world driving scenes and are the established benchmarks for comparing with previous methods \cite{liOcclusionAwareStereo2019,wangParallaxAttentionUnsupervised2022,yangUnsupervisedHierarchicalIterative2024,Flow2StereoEffectiveSelfSupervised,fanOcclusionAwareSelfSupervisedStereo2022}.  To validate the robustness of our method, we also use two additional datasets with vastly different styles: SceneFlow \cite{mayerLargeDatasetTrain2016} and Spring \cite{mehlSpringHighresolutionHighdetail2023}, with the latter being used at half resolution owing to our limitations in computational resources. For the network architectures, we select three representative models with highly distinct designs---PSMNet \cite{changPyramidStereoMatching2018}, GMStereo \cite{xu2023unifying}, and MonSter \cite{chengMonSterMarryMonodepth2025}---to verify the general applicability of our method across different backbones.

\textbf{Training Details.} We maintain consistent parameters across different datasets to demonstrate the robustness of our method. All training is conducted using two RTX 4090 GPUs in PyTorch \cite{paszke2019pytorch}. We employ the Adam optimizer \cite{2015kingma} with $\beta_1=0.9$, $\beta_2=0.999$ and the cosine annealing schedule. Following the settings in \cite{changPyramidStereoMatching2018}, we crop images to $512\times256$ during training, with a batch size of 8. Data augmentation is applied to the input images but not to the feedback images. This augmentation consists of random cropping, brightness enhancement, and contrast adjustment. All training processes start from scratch without any pre-training, ground-truth labels, or post-processing involved.

\subsection{Comparisons with State-of-the-art} \label{sub-compar}
We first present our submitted results on the KITTI online leaderboard in \cref{tab:online}. The evaluation metrics are identical to those in the online benchmarks. Specifically, ``noc'' and ``all'' denote non-occluded pixels and all pixels respectively. ``n-noc/all''  refers to the percentage of pixels exceeding a difference of ``n'' from ground truth. ``D1'' indicates the proportion of pixels where the end-point error is more than 3px and 5\%. ``EPE'' represents the average difference between estimated values and ground truth.

\begin{figure}[t]
    \centering
    \includegraphics[width=1\linewidth]{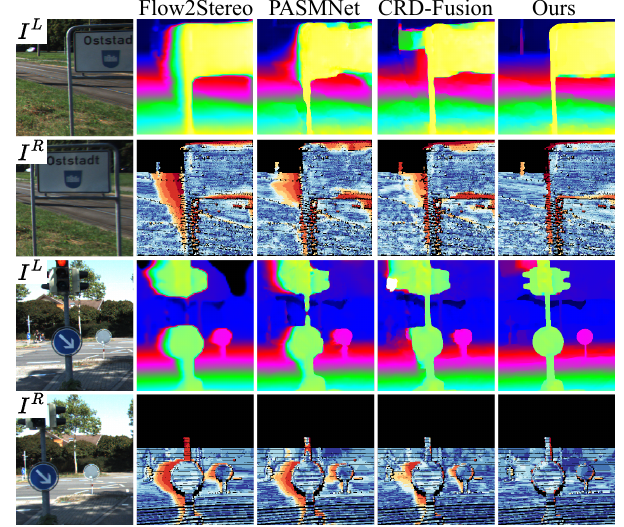}
    \caption{Qualitative comparisons. All disparity and error maps sourced from the KITTI 2015 online benchmark. Only our method produces accurate results on both sides of occluders, demonstrating that the occlusion problem is resolved.}
    \label{fig:kt}
\end{figure}

The results in \cref{tab:online} demonstrate that our method significantly outperforms other methods across the board. We present results using two distinct backbones: PSMNet \cite{changPyramidStereoMatching2018} and MonSter \cite{chengMonSterMarryMonodepth2025}. The former is chosen to ensure that the performance gains are attributable to our method rather than to advancements in network architecture, thereby validating the credibility of our contribution. The latter validates the generalizability and applicability of our method to the latest architectures.
These performance improvements are attributed to our successful resolution of the occlusion problem, which is visually demonstrated in \cref{fig:kt}. The figure presents representative occlusion scenarios alongside the disparity and error maps from different methods. It is clearly visible that other methods exhibit varying degrees of error on the left side of the occluder, which indicates that the occlusion problem remained unresolved in these prior works. In contrast, only our method produces results that are symmetric and accurate on both sides of the occluder. This validates that the primary objective of this work---to solve the occlusion challenge---has been successfully achieved. 

\subsection{Evaluation on Various Backbones and Datasets} \label{sub-more}
\begin{figure*}[t]
    \centering
    \includegraphics[width=0.95\linewidth]{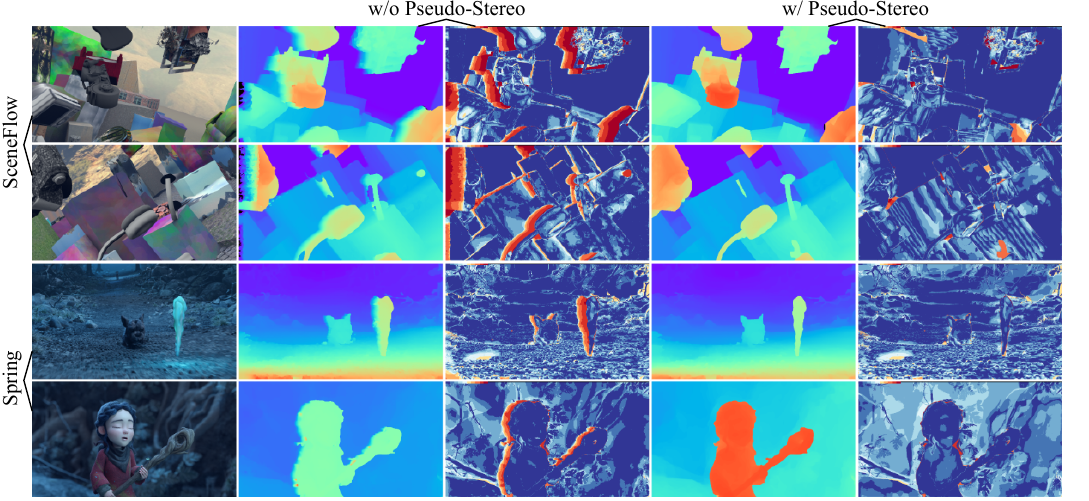}
    \caption{Qualitative results of our proposed Pseudo-Stereo method on the SceneFlow~\cite{mayerLargeDatasetTrain2016} and Spring~\cite{mehlSpringHighresolutionHighdetail2023} datasets, demonstrating its ability to resolve the occlusion challenge across various scenarios.}
    \label{fig:more_datasets}
    % \vspace{-1em}
\end{figure*}

This section aims to demonstrate the robustness and effectiveness of our approach across various backbone architectures and datasets. As shown in \cref{fig:more_datasets}, the occlusion problem is prevalent across various datasets, manifesting as erroneous disparity inference on the left side of foreground occluders (indicated by red pixels in the error maps). The proposed Pseudo-Stereo strategy demonstrates clear effectiveness in these occluded regions, yielding accurate results on both sides of the occluder. This qualitative improvement translates to quantitative gains, as shown in \cref{tab:more_datasets}, where applying our method to different backbones on various datasets all leads to substantial accuracy enhancements attributable to solving the occlusion problem.

\subsection{Ablation Study} \label{sub-ablation}

In this section, the roles of the various components proposed in \cref{sec:method} within the overall framework are studied and discussed. All evaluations are performed on the KITTI 2015 training set. The components involved include the pseudo-stereo inputs strategy, the refined pseudo-image generation method, the occlusion estimation, and the fully pseudo-stereo inputs strategy. The experimental results are shown in \cref{tab:ablation}. 

\begin{figure}[t]
    \centering
    \includegraphics[width=1\linewidth]{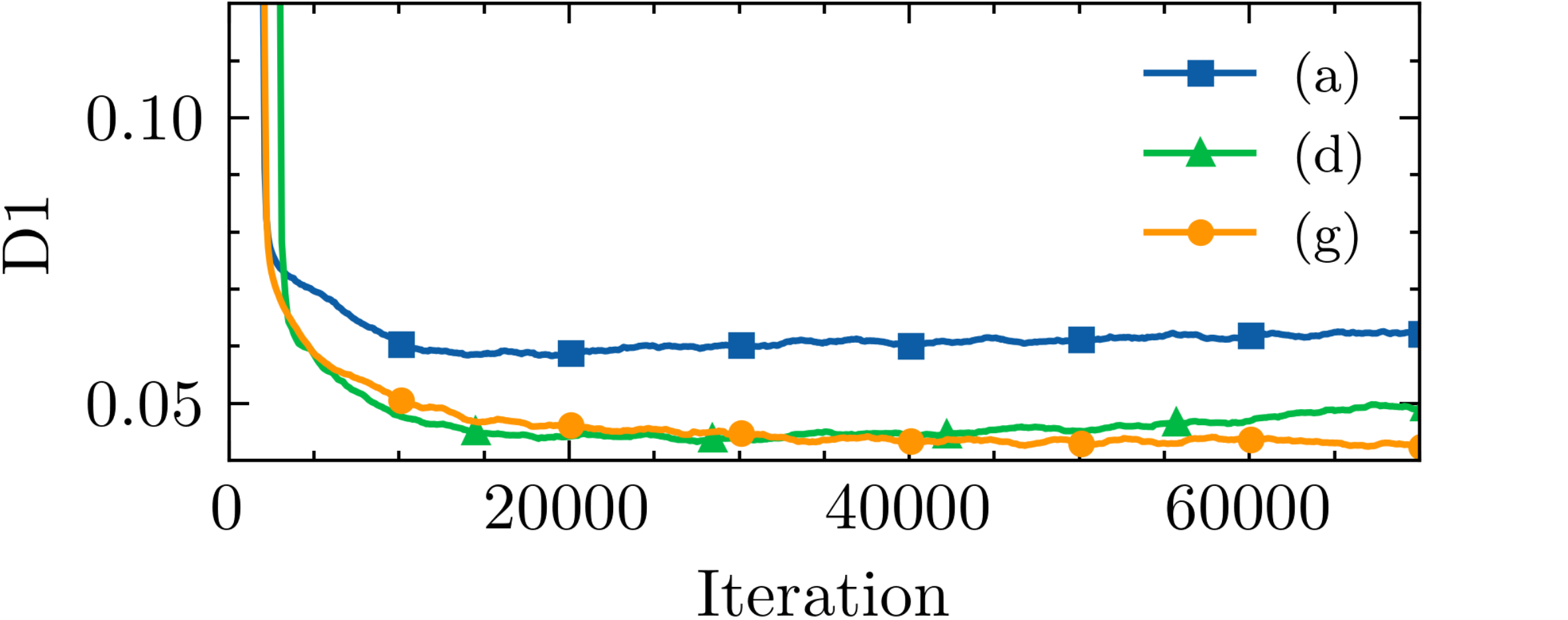}
    \caption{Visualization of the overfitting problem. By adopting the fully pseudo-stereo input strategy in (f), the increasing D1 error trend observed in (c) is eliminated.}
    \label{fig:overfitting}
    \vspace{-1em}
\end{figure}

The results for settings (a-d) indicate that both the pseudo-stereo inputs and occlusion estimation are essential for performance enhancement, but they concurrently induce overfitting. Conversely, the results from (e-g), where overfitting is absent, prove that our fully pseudo-stereo input strategy is the key to resolving the overfitting problem. 
In \cref{fig:overfitting}, we plot the training curves to visually illustrate this phenomenon. 
The comparison between (f) and (g) reveals that when the generated pseudo-image suffers from significant artifacts, adopting the fully pseudo-stereo strategy can lead to a degradation in accuracy. This issue is effectively resolved when our refined pseudo-image generation strategy is used. Overall, each component plays an indispensable role within the entire framework. The combination of all components enables our model to achieve both high accuracy and stable training.

\begin{table}[t]
\centering
\caption{Evaluation on various backbones and datasets.}
\label{tab:more_datasets}
\resizebox{\columnwidth}{!}{%
\begin{tabular}{lccccc}
\toprule
\multirow{2}{*}{Backbone}                       & \multicolumn{2}{c}{w/o Pseudo-Stereo}                   &  & \multicolumn{2}{c}{w/ Pseudo-Stereo}                       \\ \cline{2-3} \cline{5-6} 
                                                & EPE (px) $\downarrow$ & 3-px (\%) $\downarrow$ &  & EPE (px) $\downarrow$ & 3-px (\%) $\downarrow$ \\ \hline
\multicolumn{6}{c}{\textit{SceneFlow~\cite{mayerLargeDatasetTrain2016} Dataset}}                                                                                              \\
PSMNet~\cite{changPyramidStereoMatching2018}    & 5.41                  & 16.85                  &  & \textbf{3.10}             & \textbf{9.80}              \\
GMStereo~\cite{xu2023unifying}                  & 5.09                  & 15.01                  &  & \textbf{3.11}         & \textbf{8.49}          \\
MonSter~\cite{chengMonSterMarryMonodepth2025} & 5.28                  & 16.15                  &  & \textbf{2.32}         & \textbf{7.86}          \\ \hline
\multicolumn{6}{c}{\textit{Spring~\cite{mehlSpringHighresolutionHighdetail2023} Dataset}}                                                                                                 \\
PSMNet~\cite{changPyramidStereoMatching2018}    & 2.41                  & 6.58                   &  & \textbf{2.30}         & \textbf{5.12}              \\
GMStereo~\cite{xu2023unifying}                  & 2.33                  & 6.31                   &  & \textbf{2.18}         & \textbf{4.78}          \\
MonSter~\cite{chengMonSterMarryMonodepth2025} & 2.37                  & 6.56                   &  & \textbf{1.77}         & \textbf{4.65}          \\ \bottomrule
\end{tabular}%
}
\end{table}

\begin{table}[t]
\centering
\caption{Ablation study of our proposed components. PS: Pseudo-stereo inputs strategy. Occ: Occlusion estimation and gradient detaching. RG: Refined pseudo-image generation. FPS: Fully pseudo-stereo inputs strategy.}
\label{tab:ablation}
\resizebox{\columnwidth}{!}{%
\begin{tabular}{cccccccc}
\toprule
Config. & PS         & Occ        & RG         & FPS        & D1 (\%) & EPE  & Remarks                                         \\ \hline
(a)     &            &            &            &            & 5.67    & 1.17 & /                                               \\
(b)     &            & \checkmark &            &            & 5.73    & 1.24 & \multicolumn{1}{l}{Over-smoothing in occlusions} \\ \hline
(c)     & \checkmark &            &            &            & 4.99    & 1.07 & Overfitting                                     \\
(d)     & \checkmark & \checkmark &            &            & 4.08    & 0.99 & Overfitting                                     \\ \hline
(e)     & \checkmark &            &            & \checkmark & 5.17    & 1.11 & /                                               \\
(f)     & \checkmark & \checkmark &            & \checkmark & 4.31    & 1.04 & /                                               \\
(g)     & \checkmark & \checkmark & \checkmark & \checkmark & 4.06    & 1.01 & /                                               \\   \bottomrule
\end{tabular}%
}
\end{table}

\section{Conclusion}
In this work, we propose a self-supervised stereo matching paradigm with the core idea of a pseudo-stereo inputs strategy. This method effectively addresses the occlusion issue, which has consistently hindered the performance of self-supervised stereo matching. Extensive experiments validate our method's success in handling occlusions and demonstrate its robustness and generality across diverse datasets and network backbones.

Despite our significant progress on occlusions, this work does not propose explicit solutions for other challenges, such as textureless and reflective regions. In fact, unlike occlusions, providing accurate feedback for these ill-posed regions is inherently intractable with binocular cues alone. A promising solution is the multi-frame self-supervised paradigm, as explored in \cite{wangUnifiedUnsupervisedOpticalFlow2019a}. This creates a powerful synergy: temporal consistency across frames is key to resolving ambiguities in areas like textureless regions, while the accuracy of the single-frame disparity estimation is core to resolving depth ambiguities of moving objects. The substantially enhanced reliability of disparity estimation achieved by our work provides a much stronger foundation for this paradigm, creating significant potential to better leverage multi-frame information and tackle the remaining challenges in self-supervised 3D perception.

{
    \small
    \bibliographystyle{ieeenat_fullname}
    \bibliography{main.bib}

\begin{thebibliography}{37}
\providecommand{\natexlab}[1]{#1}
\providecommand{\url}[1]{\texttt{#1}}
\expandafter\ifx\csname urlstyle\endcsname\relax
  \providecommand{\doi}[1]{doi: #1}\else
  \providecommand{\doi}{doi: \begingroup \urlstyle{rm}\Url}\fi

\bibitem[Chang and Chen(2018)]{changPyramidStereoMatching2018}
Jia-Ren Chang and Yong-Sheng Chen.
\newblock Pyramid stereo matching network.
\newblock In \emph{Proceedings of the {{IEEE}} Conference on Computer Vision
  and Pattern Recognition}, pages 5410--5418, 2018.

\bibitem[Cheng et~al.(2025)Cheng, Liu, Xu, Wang, Zhang, Deng, Zang, Chen, Cai,
  and Yang]{chengMonSterMarryMonodepth2025}
Junda Cheng, Longliang Liu, Gangwei Xu, Xianqi Wang, Zhaoxing Zhang, Yong Deng,
  Jinliang Zang, Yurui Chen, Zhipeng Cai, and Xin Yang.
\newblock {{MonSter}}: {{Marry Monodepth}} to {{Stereo Unleashes Power}}.
\newblock In \emph{Proceedings of the {{IEEE}}/{{CVF Conference}} on {{Computer
  Vision}} and {{Pattern Recognition}} ({{CVPR}})}, pages 6273--6282, 2025.

\bibitem[Fan et~al.(2022)Fan, Jeon, and
  Fidan]{fanOcclusionAwareSelfSupervisedStereo2022}
Xiule Fan, Soo Jeon, and Baris Fidan.
\newblock Occlusion-{{Aware Self-Supervised Stereo Matching}} with {{Confidence
  Guided Raw Disparity Fusion}}.
\newblock In \emph{2022 19th {{Conference}} on {{Robots}} and {{Vision}}
  ({{CRV}})}, pages 132--139, 2022.

\bibitem[Fang et~al.(2023)Fang, Wen, Hsu, Jen, Chen, Chen,
  et~al.]{fangESNetAccurate2023}
I Fang, Hsiao-Chieh Wen, Chia-Lun Hsu, Po-Chung Jen, Ping-Yang Chen, Yong-Sheng
  Chen, et~al.
\newblock {{ES3Net}}: {{Accurate}} and efficient edge-based self-supervised
  stereo matching network.
\newblock In \emph{Proceedings of the {{IEEE}}/{{CVF}} Conference on Computer
  Vision and Pattern Recognition}, pages 4472--4481, 2023.

\bibitem[Geiger et~al.(2012)Geiger, Lenz, and Urtasun]{geigerAreWeReady2012}
A. Geiger, P. Lenz, and R. Urtasun.
\newblock Are we ready for autonomous driving? {{The KITTI}} vision benchmark
  suite.
\newblock In \emph{2012 {{IEEE Conference}} on {{Computer Vision}} and
  {{Pattern Recognition}}}, pages 3354--3361, Providence, RI, 2012. IEEE.

\bibitem[Godard et~al.(2017)Godard, Aodha, and
  Brostow]{godardUnsupervisedMonocularDepth2017}
Clement Godard, Oisin~Mac Aodha, and Gabriel~J. Brostow.
\newblock Unsupervised {{Monocular Depth Estimation}} with {{Left-Right
  Consistency}}.
\newblock In \emph{2017 {{IEEE Conference}} on {{Computer Vision}} and
  {{Pattern Recognition}} ({{CVPR}})}, pages 6602--6611, Honolulu, HI, 2017.
  IEEE.

\bibitem[Godard et~al.(2019)Godard, Aodha, Firman, and
  Brostow]{godardDiggingSelfSupervisedMonocular2019}
Clement Godard, Oisin~Mac Aodha, Michael Firman, and Gabriel Brostow.
\newblock Digging {{Into Self-Supervised Monocular Depth Estimation}}.
\newblock In \emph{2019 {{IEEE}}/{{CVF International Conference}} on {{Computer
  Vision}} ({{ICCV}})}, pages 3827--3837, Seoul, Korea (South), 2019. IEEE.

\bibitem[Hamid et~al.(2022)Hamid, Abd~Manap, Hamzah, and
  Kadmin]{hamidStereoMatchingAlgorithm2022}
Mohd~Saad Hamid, NurulFajar Abd~Manap, Rostam~Affendi Hamzah, and Ahmad~Fauzan
  Kadmin.
\newblock Stereo matching algorithm based on deep learning: {{A}} survey.
\newblock \emph{Journal of King Saud University-Computer and Information
  Sciences}, 34\penalty0 (5):\penalty0 1663--1673, 2022.

\bibitem[Hirschmuller(2005)]{hirschmullerAccurateEfficientStereo2005}
H. Hirschmuller.
\newblock Accurate and efficient stereo processing by semi-global matching and
  mutual information.
\newblock In \emph{2005 {{IEEE Computer Society Conference}} on {{Computer
  Vision}} and {{Pattern Recognition}} ({{CVPR}}'05)}, pages 807--814 vol. 2,
  2005.

\bibitem[Hur and Roth(2021)]{hurSelfSupervisedMultiFrameMonocular2021}
Junhwa Hur and Stefan Roth.
\newblock Self-{{Supervised Multi-Frame Monocular Scene Flow}}.
\newblock In \emph{2021 {{IEEE}}/{{CVF Conference}} on {{Computer Vision}} and
  {{Pattern Recognition}} ({{CVPR}})}, pages 2683--2693, Nashville, TN, USA,
  2021. IEEE.

\bibitem[Janai et~al.(2018)Janai, Guney, Ranjan, Black, and
  Geiger]{Janai_2018_ECCV}
Joel Janai, Fatma Guney, Anurag Ranjan, Michael Black, and Andreas Geiger.
\newblock Unsupervised learning of multi-frame optical flow with occlusions.
\newblock In \emph{Proceedings of the {{European Conference}} on {{Computer
  Vision}} ({{ECCV}})}, 2018.

\bibitem[Jonschkowski et~al.(2020)Jonschkowski, Stone, Barron, Gordon,
  Konolige, and Angelova]{jonschkowski2020matters}
Rico Jonschkowski, Austin Stone, Jonathan~T Barron, Ariel Gordon, Kurt
  Konolige, and Anelia Angelova.
\newblock What matters in unsupervised optical flow.
\newblock In \emph{Computer {{Vision}}--{{ECCV}} 2020: 16th European
  Conference, Glasgow, {{UK}}, August 23--28, 2020, Proceedings, Part {{II}}
  16}, pages 557--572. Springer, 2020.

\bibitem[Kerbl et~al.(2023)Kerbl, Kopanas, Leimkuehler, and
  Drettakis]{kerbl3DGaussianSplatting2023}
Bernhard Kerbl, Georgios Kopanas, Thomas Leimkuehler, and George Drettakis.
\newblock {{3D Gaussian Splatting}} for {{Real-Time Radiance Field Rendering}}.
\newblock \emph{ACM Transactions on Graphics}, 42\penalty0 (4):\penalty0 1--14,
  2023.

\bibitem[Khan et~al.(2023)Khan, Xiao, and
  Lanman]{khanTiledMultiplaneImages2023}
Numair Khan, Lei Xiao, and Douglas Lanman.
\newblock Tiled {{Multiplane Images}} for {{Practical 3D Photography}}.
\newblock In \emph{2023 {{IEEE}}/{{CVF International Conference}} on {{Computer
  Vision}} ({{ICCV}})}, pages 10420--10430, Paris, France, 2023. IEEE.

\bibitem[Kingma and Ba(2015)]{2015kingma}
Diederik~P. Kingma and Jimmy Ba.
\newblock Adam: {{A}} method for stochastic optimization.
\newblock In \emph{The {{Eleventh International Conference}} on {{Learning
  Representations}}}, 2015.

\bibitem[Li and Yuan(2018)]{liOcclusionAwareStereo2019}
Ang Li and Zejian Yuan.
\newblock Occlusion aware stereo matching via cooperative unsupervised
  learning.
\newblock In \emph{Asian Conference on Computer Vision}, pages 197--213.
  Springer, 2018.

\bibitem[Li et~al.(2022)Li, Yuan, Ling, Chi, Zhang, and
  Zhang]{liUnsupervisedOcclusionawareStereo2022}
Ang Li, Zejian Yuan, Yonggen Ling, Wanchao Chi, Shenghao Zhang, and Chong
  Zhang.
\newblock Unsupervised occlusion-aware stereo matching with directed disparity
  smoothing.
\newblock \emph{IEEE Transactions on Intelligent Transportation Systems},
  23\penalty0 (7):\penalty0 7457--7468, 2022.

\bibitem[Li et~al.(2024)Li, Wang, Zhang, Xue, Zhou, and Guo]{li2024local}
Kunhong Li, Longguang Wang, Ye Zhang, Kaiwen Xue, Shunbo Zhou, and Yulan Guo.
\newblock {{LoS}}: {{Local}} structure-guided stereo matching.
\newblock In \emph{Proceedings of the {{IEEE}}/{{CVF}} Conference on Computer
  Vision and Pattern Recognition}, pages 19746--19756, 2024.

\bibitem[Liu et~al.(2020)Liu, King, Lyu, and
  Xu]{Flow2StereoEffectiveSelfSupervised}
Pengpeng Liu, Irwin King, Michael~R. Lyu, and Jia Xu.
\newblock {{Flow2Stereo}}: {{Effective Self-Supervised Learning}} of {{Optical
  Flow}} and {{Stereo Matching}}.
\newblock In \emph{2020 {{IEEE}}/{{CVF Conference}} on {{Computer Vision}} and
  {{Pattern Recognition}} ({{CVPR}})}, pages 6647--6656, 2020.

\bibitem[Mayer et~al.(2016)Mayer, Ilg, Hausser, Fischer, Cremers, Dosovitskiy,
  and Brox]{mayerLargeDatasetTrain2016}
Nikolaus Mayer, Eddy Ilg, Philip Hausser, Philipp Fischer, Daniel Cremers,
  Alexey Dosovitskiy, and Thomas Brox.
\newblock A {{Large Dataset}} to {{Train Convolutional Networks}} for
  {{Disparity}}, {{Optical Flow}}, and {{Scene Flow Estimation}}.
\newblock In \emph{2016 {{IEEE Conference}} on {{Computer Vision}} and
  {{Pattern Recognition}} ({{CVPR}})}, pages 4040--4048, Las Vegas, NV, USA,
  2016. IEEE.

\bibitem[Mehl et~al.(2023)Mehl, Schmalfuss, Jahedi, Nalivayko, and
  Bruhn]{mehlSpringHighresolutionHighdetail2023}
Lukas Mehl, Jenny Schmalfuss, Azin Jahedi, Yaroslava Nalivayko, and Andr{\'e}s
  Bruhn.
\newblock Spring: A high-resolution high-detail dataset and benchmark for scene
  flow, optical flow and stereo.
\newblock In \emph{2023 {{IEEE}}/{{CVF Conference}} on {{Computer Vision}} and
  {{Pattern Recognition}} ({{CVPR}})}, pages 4981--4991, Vancouver, BC, Canada,
  2023. IEEE.

\bibitem[Meister et~al.(2018)Meister, Hur, and Roth]{Meister:2018:UUL}
Simon Meister, Junhwa Hur, and Stefan Roth.
\newblock {{UnFlow}}: Unsupervised learning of optical flow with a
  bidirectional census loss.
\newblock In \emph{Aaai}, New Orleans, Louisiana, 2018.

\bibitem[Menze and Geiger(2015)]{menzeObjectSceneFlow2015}
Moritz Menze and Andreas Geiger.
\newblock Object {{Scene Flow}} for {{Autonomous Vehicles}}.
\newblock In \emph{Proceedings of the {{IEEE Conference}} on {{Computer
  Vision}} and {{Pattern Recognition}}}, pages 3061--3070, 2015.

\bibitem[Mildenhall et~al.(2020)Mildenhall, Srinivasan, Tancik, Barron,
  Ramamoorthi, and Ng]{mildenhall2020nerf}
Ben Mildenhall, Pratul~P Srinivasan, Matthew Tancik, Jonathan~T Barron, Ravi
  Ramamoorthi, and Ren Ng.
\newblock {{NeRF}}: Representing scenes as neural radiance fields for view
  synthesis.
\newblock In \emph{European {{Conference}} on {{Computer Vision}} ({{ECCV}})},
  pages 405--421. Springer, 2020.

\bibitem[Paszke et~al.(2019)Paszke, Gross, Massa, Lerer, Bradbury, Chanan,
  Killeen, Lin, Gimelshein, Antiga, et~al.]{paszke2019pytorch}
Adam Paszke, Sam Gross, Francisco Massa, Adam Lerer, James Bradbury, Gregory
  Chanan, Trevor Killeen, Zeming Lin, Natalia Gimelshein, Luca Antiga, et~al.
\newblock Pytorch: {{An}} imperative style, high-performance deep learning
  library.
\newblock \emph{Advances in neural information processing systems}, 32, 2019.

\bibitem[Poggi et~al.(2021)Poggi, Tosi, Batsos, Mordohai, and
  Mattoccia]{poggiSynergiesMachineLearning2021}
Matteo Poggi, Fabio Tosi, Konstantinos Batsos, Philippos Mordohai, and Stefano
  Mattoccia.
\newblock On the {{Synergies}} between {{Machine Learning}} and {{Binocular
  Stereo}} for {{Depth Estimation}} from {{Images}}: A {{Survey}}.
\newblock \emph{IEEE Transactions on Pattern Analysis and Machine
  Intelligence}, pages 1--1, 2021.

\bibitem[Tankovich et~al.(2021)Tankovich, Hane, Zhang, Kowdle, Fanello, and
  Bouaziz]{tankovichHITNetHierarchicalIterative2021}
Vladimir Tankovich, Christian Hane, Yinda Zhang, Adarsh Kowdle, Sean Fanello,
  and Sofien Bouaziz.
\newblock {{HITNet}}: {{Hierarchical Iterative Tile Refinement Network}} for
  {{Real-time Stereo Matching}}.
\newblock In \emph{2021 {{Ieee}}/{{Cvf Conference}} on {{Computer Vision}} and
  {{Pattern Recognition}}, {{Cvpr}} 2021}, pages 14357--14367, 2021.

\bibitem[Tosi et~al.(2023)Tosi, Tonioni, De~Gregorio, and
  Poggi]{tosiNeRFSupervisedDeepStereo2023}
Fabio Tosi, Alessio Tonioni, Daniele De~Gregorio, and Matteo Poggi.
\newblock {{NeRF-Supervised Deep Stereo}}.
\newblock In \emph{2023 {{IEEE}}/{{CVF Conference}} on {{Computer Vision}} and
  {{Pattern Recognition}} ({{CVPR}})}, pages 855--866, Vancouver, BC, Canada,
  2023. IEEE.

\bibitem[Tosi et~al.(2025)Tosi, Bartolomei, and
  Poggi]{tosiSurveyDeepStereo2025}
Fabio Tosi, Luca Bartolomei, and Matteo Poggi.
\newblock A {{Survey}} on {{Deep Stereo Matching}} in the {{Twenties}}.
\newblock \emph{International Journal of Computer Vision}, 2025.

\bibitem[Vaswani et~al.(2017)Vaswani, Shazeer, Parmar, Uszkoreit, Jones, Gomez,
  Kaiser, and Polosukhin]{transformer}
Ashish Vaswani, Noam Shazeer, Niki Parmar, Jakob Uszkoreit, Llion Jones,
  Aidan~N. Gomez, {\L}ukasz Kaiser, and Illia Polosukhin.
\newblock Attention is all you need.
\newblock In \emph{Proceedings of the 31st International Conference on Neural
  Information Processing Systems}, pages 6000--6010, Red Hook, NY, USA, 2017.
  Curran Associates Inc.

\bibitem[Wang et~al.(2020)Wang, Guo, Wang, Liang, Lin, Yang, and
  An]{wangParallaxAttentionUnsupervised2022}
Longguang Wang, Yulan Guo, Yingqian Wang, Zhengfa Liang, Zaiping Lin, Jungang
  Yang, and Wei An.
\newblock Parallax attention for unsupervised stereo correspondence learning.
\newblock \emph{IEEE transactions on pattern analysis and machine
  intelligence}, 44\penalty0 (4):\penalty0 2108--2125, 2020.

\bibitem[Wang et~al.(2018)Wang, Yang, Yang, Zhao, Wang, and
  Xu]{wangOcclusionAwareUnsupervised2018}
Yang Wang, Yi Yang, Zhenheng Yang, Liang Zhao, Peng Wang, and Wei Xu.
\newblock Occlusion {{Aware Unsupervised Learning}} of {{Optical Flow}}.
\newblock In \emph{2018 {{IEEE}}/{{CVF Conference}} on {{Computer Vision}} and
  {{Pattern Recognition}}}, pages 4884--4893, Salt Lake City, UT, 2018. IEEE.

\bibitem[Wang et~al.(2019)Wang, Wang, Yang, Luo, Yang, and
  Xu]{wangUnifiedUnsupervisedOpticalFlow2019a}
Yang Wang, Peng Wang, Zhenheng Yang, Chenxu Luo, Yi Yang, and Wei Xu.
\newblock {{UnOS}}: {{Unified Unsupervised Optical-Flow}} and {{Stereo-Depth
  Estimation}} by {{Watching Videos}}.
\newblock In \emph{2019 {{IEEE}}/{{CVF Conference}} on {{Computer Vision}} and
  {{Pattern Recognition}} ({{CVPR}})}, pages 8063--8073, Long Beach, CA, USA,
  2019. IEEE.

\bibitem[Xu et~al.(2023)Xu, Zhang, Cai, Rezatofighi, Yu, Tao, and
  Geiger]{xu2023unifying}
Haofei Xu, Jing Zhang, Jianfei Cai, Hamid Rezatofighi, Fisher Yu, Dacheng Tao,
  and Andreas Geiger.
\newblock Unifying flow, stereo and depth estimation.
\newblock \emph{IEEE Transactions on Pattern Analysis and Machine
  Intelligence}, 2023.

\bibitem[Yang et~al.(2024)Yang, Li, Cong, and
  Du]{yangUnsupervisedHierarchicalIterative2024}
Ruizhi Yang, Xingqiang Li, Rigang Cong, and Jinsong Du.
\newblock Unsupervised {{Hierarchical Iterative Tile Refinement Network With 3D
  Planar Segmentation Loss}}.
\newblock \emph{IEEE Robotics and Automation Letters}, 9\penalty0 (3):\penalty0
  2678--2685, 2024.

\bibitem[Yuan et~al.(2021)Yuan, Zhang, Wu, Zhu, Tan, Wang, and
  Chen]{yuanStereoMatchingSelfsupervision2021}
Weihao Yuan, Yazhan Zhang, Bingkun Wu, Siyu Zhu, Ping Tan, Michael~Yu Wang, and
  Qifeng Chen.
\newblock Stereo matching by self-supervision of multiscopic vision, 2021.

\bibitem[Zhou et~al.(2017)Zhou, Zhang, Shen, and
  Jia]{zhouUnsupervisedLearningStereo2017}
Chao Zhou, Hong Zhang, Xiaoyong Shen, and Jiaya Jia.
\newblock Unsupervised {{Learning}} of {{Stereo Matching}}.
\newblock In \emph{2017 {{IEEE International Conference}} on {{Computer
  Vision}} ({{ICCV}})}, pages 1576--1584, Venice, 2017. IEEE.

\end{thebibliography}
}


\end{document}